\documentclass[runningheads]{llncs}
\usepackage[T1]{fontenc}
\usepackage{graphicx}
\usepackage{amssymb,amsfonts}
\usepackage{algorithmic}
\usepackage{textcomp}
\usepackage{xcolor}
\usepackage{float}
\usepackage{caption}
\usepackage{subfig}
\usepackage{amsmath}
\usepackage{hyperref}
\usepackage{color}

\usepackage{booktabs}
\usepackage{orcidlink}

\begin{document}

\title{Stroke Locus Net: Occluded Vessel Localization from MRI Modalities}

\titlerunning{Stroke Locus Net}

\author{Mohamed Hamad\inst{1}
\and
Muhammad Khan\inst{1} \and
Tamer Khattab\inst{1} \and
Mohamed Mabrok\inst{2}
}

\authorrunning{M. Hamad et al.}

\institute{Department of Electrical Engineering, College of Engineering, Qatar University, Doha, Qatar
\email{\{m.bashir.hamad, mm2105963, tkhattab\}@qu.edu.qa}
\and
Department of Mathematics and Statistics, College of Arts and Sciences, Qatar University, Doha, Qatar
\email{m.a.mabrok@qu.edu.qa}}

\maketitle

\begin{abstract}
A key challenge in ischemic stroke diagnosis using medical imaging is the accurate localization of the occluded vessel. Current machine learning methods in focus primarily on lesion segmentation, with limited work on vessel localization. In this study, we introduce Stroke Locus Net, an end-to-end deep learning pipeline for detection, segmentation, and occluded vessel localization using only MRI scans. The proposed system combines a segmentation branch using nnUNet for lesion detection with an arterial atlas for vessel mapping and identification, and a generation branch using pGAN to synthesize MRA images from MRI. Our implementation demonstrates promising results in localizing occluded vessels on stroke-affected T1 MRI scans, with potential for faster and more informed stroke diagnosis. \footnote{AI tools such as ChatGPT and Claude AI have been used in various sections of this work for phrasing and sentence structure.}

\end{abstract}

\section{Introduction}
Ischemic stroke remains one of the leading causes of death and disability worldwide \cite{patil2022leading}. Stroke occurs when a blood clot blocks cerebral blood flow, causing ischemia, a deficiency of oxygen and nutrients in the affected brain tissue. Consequently, this ischemic tissue may undergo infarction, leading to cell death and the formation of a "lesion" \cite{MedlinePlusStroke}. The lesion, therefore, represents the area of damaged brain tissue resulting from the vascular event.

Lesion segmentation, the process of identifying and delineating the boundaries of the stroke-induced lesion in medical images, allows clinicians to identify the occluded (stroke-causing) vessel. This is crucial for determining the extent of brain damage, appropriate treatment strategy, and predicting patient outcomes \cite{stebner2025lvo-medium-small}. It also helps decide whether interventions like thrombolysis or endovascular thrombectomy are necessary. Furthermore, speed in identifying the occluded vessel is important, particularly in the acute phase \cite{2020strokemanagement}. The primary reason for this is that the immediate treatment decisions, especially those aimed at restoring blood flow, hinge on identifying and targeting the occluded vessel \cite{stebner2025lvo-medium-small}.

While medical imaging techniques such as MRI and CT (Computed Tomography) are essential for stroke diagnosis, standard MRI sequences like T1-weighted or T2-weighted images primarily depict brain parenchyma and are not optimized for visualizing blood vessels \cite{nukovic2023neuroimaging}. In contrast, Magnetic Resonance Angiography (MRA) is specifically designed to capture blood vessels, providing detailed anatomical and functional information about the cerebrovascular system without using ionizing radiation. However, MRA sequences are not always included in standard acute stroke MRI protocols due to factors like scan time and cost \cite{olut2018sequences}. CT angiography (CTA) is another modality used for vessel imaging in stroke, offering rapid acquisition and high accuracy in detecting large vessel occlusions. However, CTA involves ionizing radiation and requires the administration of iodinated contrast agents, which may not be suitable for all patients \cite{nukovic2023neuroimaging}.

Advances in deep learning have shown promising results in medical image analysis, particularly in stroke lesion segmentation \cite{isensee2021nnunet,ronneberger2015unet}. However, most existing approaches focus solely on lesion detection and segmentation \cite{maier2017ISLES}, with limited attention to vessel localization. The ISLES challenge \cite{maier2017ISLES} has been instrumental in advancing stroke lesion segmentation, but there remains a significant gap in addressing the critical task of vessel occlusion detection.

The main contribution of this paper is the development of an end-to-end deep learning pipeline for stroke diagnosis that combines lesion segmentation with vessel localization. We use arterial atlases to determine stroke source \cite{liu2023probabilistic} based on the stroke lesion segmented using nnUNet \cite{isensee2021nnunet}. We implement a GAN-based system \cite{goodfellow2014generative} for MRA synthesis from MRI scans and implement brain vessel segmentation methods \cite{Sabrowsky2023segmentation} to localize stroke source based on the arterial atlas, and validate the proposed method using the ATLAS dataset \cite{liew2022atlas}.

\section{Related Work}
\subsection{Stroke Lesion Segmentation}
Stroke lesion segmentation, from a medical perspective, involves visually identifying and outlining the area of brain tissue that has been damaged as a result of a stroke. This process provides crucial anatomical information about the location and extent of the infarct. From a computer vision standpoint, stroke lesion segmentation is typically framed as a semantic segmentation task where the goal is to classify each voxel (3D pixel) in a medical image as either belonging to the lesion or the healthy tissue \cite{malik2024stroke,nouman2024neuro}. This is achieved by training deep learning models on a dataset of medical images (usually MRI) where the lesions have been manually annotated by medical professionals, creating binary masks that serve as the ground truth. The trained model then learns to predict these segmentation masks for new, unseen stroke images.

Several publicly available datasets have been instrumental in advancing the field of automated stroke lesion segmentation. The Ischemic Stroke Lesion Segmentation (ISLES) challenge, with its various iterations (e.g., ISLES 2015, ISLES 2022) \cite{maier2017ISLES,mabrok2024challenges}, provides multi-center MRI datasets with diverse stroke lesion characteristics. The Anatomical Tracings of Lesion After Stroke (ATLAS) dataset \cite{liew2022atlas} is another significant resource, offering a large-scale collection of T1-weighted MRI scans with manually segmented lesions.

Recent years have witnessed significant progress in stroke lesion segmentation, largely driven by deep learning models. One of the foundational architectures in this domain is the U-Net, introduced in 2015 by Ronneberger et al. \cite{soliman2024deep,ronneberger2015unet}, which is proven to be effective for image segmentation. Building upon the success of U-Net, nnU-Net ("no new net") is a self-configuring method for deep learning-based biomedical image segmentation developed by Isensee et al. \cite{isensee2021nnunet}. Its main philosophy is to automatically configure the entire segmentation pipeline, including preprocessing, network architecture, training, and post-processing, for any new task without manual intervention. nnU-Net achieves this by employing a set of fixed parameters, interdependent rules based on dataset properties, and empirical decisions learned from a large and diverse pool of biomedical image datasets. This approach has enabled nnU-Net to surpass most existing methods, including highly specialized solutions, across a wide range of biomedical segmentation challenges, establishing it as a state-of-the-art method and a strong baseline in the field.

While nnU-Net remains a dominant force, other models are continuously being developed to push the boundaries of stroke lesion segmentation. For instance, attention mechanisms have been integrated into U-Net architectures (Attention U-Net) to focus on relevant image regions and enhance segmentation accuracy \cite{mdpi2024enhanced}. Furthermore, hybrid architectures combining convolutional networks with transformers, like the DAE-Former and LKA/DLKA networks, have shown promising results on datasets like ISLES 2022 and ATLAS v2.0, although nnU-Net often still achieves superior performance \cite{arxiv2024brain,zafari2024transformers}. 

\subsection{Vessel Localization Methods}
Vessel localization within medical imaging is an active and evolving field, while perhaps receiving comparatively less attention than stroke lesion segmentation. MRI-based vessel localization encompasses several techniques. Vessel Wall MRI (VW-MRI) is increasingly used to directly visualize intracranial artery walls and their pathological alterations, aiding in the differentiation of atherosclerotic and non-atherosclerotic vasculopathies. Techniques like Time-of-Flight (TOF) MRA and contrast-enhanced MRA are commonly employed to probe the 3D spatial architecture of arteries and veins \cite{nukovic2023neuroimaging}. Furthermore, researchers are exploring methods to segment vessels directly from standard MRI sequences using deep learning approaches. For instance, Hippe et al. \cite{hippe2020confidence} developed a confidence weighting method for robust automated measurements of popliteal vessel walls in MRI. Chen et al. \cite{arxiv2019automated} proposed an automated artery localization and vessel wall segmentation system using deep learning.

CTA is another widely used modality for vessel localization, particularly for detecting large vessel occlusions caused by strokes. Automated analysis of intracranial vessels from CTA images using deep learning is an active area of research. For example, Peng et al. \cite{Peng_2024} developed deep learning models based on four-dimensional CTA for automatic detection of large vessel occlusions. Stib et al. \cite{Stib_2020} developed a deep learning model to detect large vessel occlusions at multiphase CT angiography. Research also focuses on centerline extraction from CTA images \cite{liu2024centerline}, vessel-based registration of CTA images to spatial anatomy \cite{Tangen_2024}, and unsupervised deep learning approaches for vessel segmentation in brain CTA \cite{van_Voorst_2025}. Multimodal fusion imaging techniques incorporating CTA and CTV are also being explored for refined vessel localization and surgical planning in neurovascular compression syndromes \cite{Hou_2024}.

For vessel localization, we leverage the work of Sabrowsky and Hirsch et al. \cite{Sabrowsky2023segmentation}, who applied the nnU-Net framework to segment abdominal blood vessels in contrast-enhanced CT scans. Their model was trained on a custom dataset with manual annotations for a limited set of major vessels, including the aorta and some of its branches. The method demonstrated strong performance, particularly for larger vessels, as reflected in high Dice scores. However, a key limitation of their approach lies in the restricted number of annotated vessel classes. This contrasts with anatomical arterial atlases, which often define a more extensive set of vascular structures. As a result, such models trained on a limited-class dataset may not fully capture the vascular detail required for this application and serve as the main limiting factor to the level of detail that can be reached.

\subsection{Medical Image Generation}
In this paper, we utilize medical image generation techniques to synthesize Magnetic Resonance Angiography (MRA) images from standard Magnetic Resonance Imaging (MRI) scans. Current work in medical image generation is rapidly advancing, with Generative Adversarial Networks (GANs) being a dominant methodology. Some approaches involve unpaired multi-contrast synthesis, where GANs are used to translate an input MRI image (e.g., T1-weighted) into multiple other contrasts, including MRA, without the need for paired training data \cite{Han_2024}. 

The IXI dataset \cite{ixi_dataset} is a publicly available dataset containing nearly 600 MRI scans from normal, healthy subjects. It provides a comprehensive collection of multi-modal MRI data, including T1, T2, and PD-weighted images, MRA images, and diffusion-weighted images (15 directions). This rich multi-modal dataset has been widely used in medical image generation research, particularly for learning mappings between different MRI contrasts. However, a significant limitation of using IXI for stroke-related applications is that it only contains scans from healthy individuals, without any stroke patients. This limitation could affect the model's ability to accurately synthesize MRA images from stroke patients' MRI scans, where vascular occlusions and collateral circulation patterns might differ significantly from healthy subjects.

The paper we are using as a foundation for our image generation branch uses pGANs (pixel-wise loss GAN) as the MRA generation model \cite{Dar2019MRISynthesis}. Similar to traditional GANs, pGAN leverages the generator and discriminator models, training them to generate fakes and identify them, respectively. The innovation of pGAN is the leveraging of two new loss terms in the objective function that take pixel-wise and feature map losses between the generated image and real image. One limitation of this is that the model becomes spatially dependent with the assumption that both the source and objective modalities in the dataset are perfectly registered justifying the use of pixel-wise calculations. This model also captures inter-slice dependencies by introducing the $k$ term as outlined in the paper.

\subsection{Arterial Atlases in Stroke Diagnosis}
An arterial atlas is a standardized spatial map of the brain's arterial vasculature, providing a reference for the typical location and anatomical variability of cerebral arteries within a population. These atlases serve as valuable tools in neuroimaging research and clinical practice, particularly in the context of stroke diagnosis \cite{Nowinski_2020}.

Arterial atlases are typically constructed by analyzing a large collection of vascular images, such as Magnetic Resonance Angiography (MRA) or Computed Tomography Angiography (CTA) scans, acquired from a cohort of individuals. These images undergo a process of spatial normalization, where they are aligned to a common reference brain space (e.g., Montreal Neurological Institute - MNI space) to account for individual differences in brain size and shape. For probabilistic arterial atlases, vascular territories are statistically derived across a population, and each voxel is assigned to the arterial region most likely to supply it, based on the generation process of the atlas \cite{Nowinski_2020}.

Current prominent work on arterial atlases involves their integration into computer-aided diagnosis (CAD) systems for stroke. These systems can leverage arterial atlases to automatically detect and localize vascular abnormalities indicative of stroke, such as vessel occlusions or stenoses, by comparing a patient's vascular imaging to the atlas \cite{Nowinski_2020}. Atlases can also be used to map the affected brain regions based on the known vascular supply territories of the occluded vessel, aiding in understanding the potential neurological deficits and informing treatment planning.

Liu et al. \cite{liu2023probabilistic} introduce a novel digital, three-dimensional, and deformable arterial atlas derived from a large dataset of stroke patients. Their main contribution lies in the creation of a probabilistic arterial territory atlas, where lesion masks from diffusion-weighted MR images are manually traced, transformed into a common space, and then averaged to generate probability maps for each vascular region. In addition to the simple averaging method, they propose a more advanced approach using a Bernoulli mixture model (BMM) combined with an Expectation-Maximization algorithm to capture voxel-level correlations and improve the accuracy of the probability estimates. As a result, two atlases are produced: one comprehensive atlas with 30 classes, including detailed sub-territories and ventricles, and a simplified 10-class atlas that groups these into 4 major arterial territories (ACA, MCA, PCA, and VB).

\section{Methodology}

\begin{figure}
    \centering
    \includegraphics[width=1\linewidth]{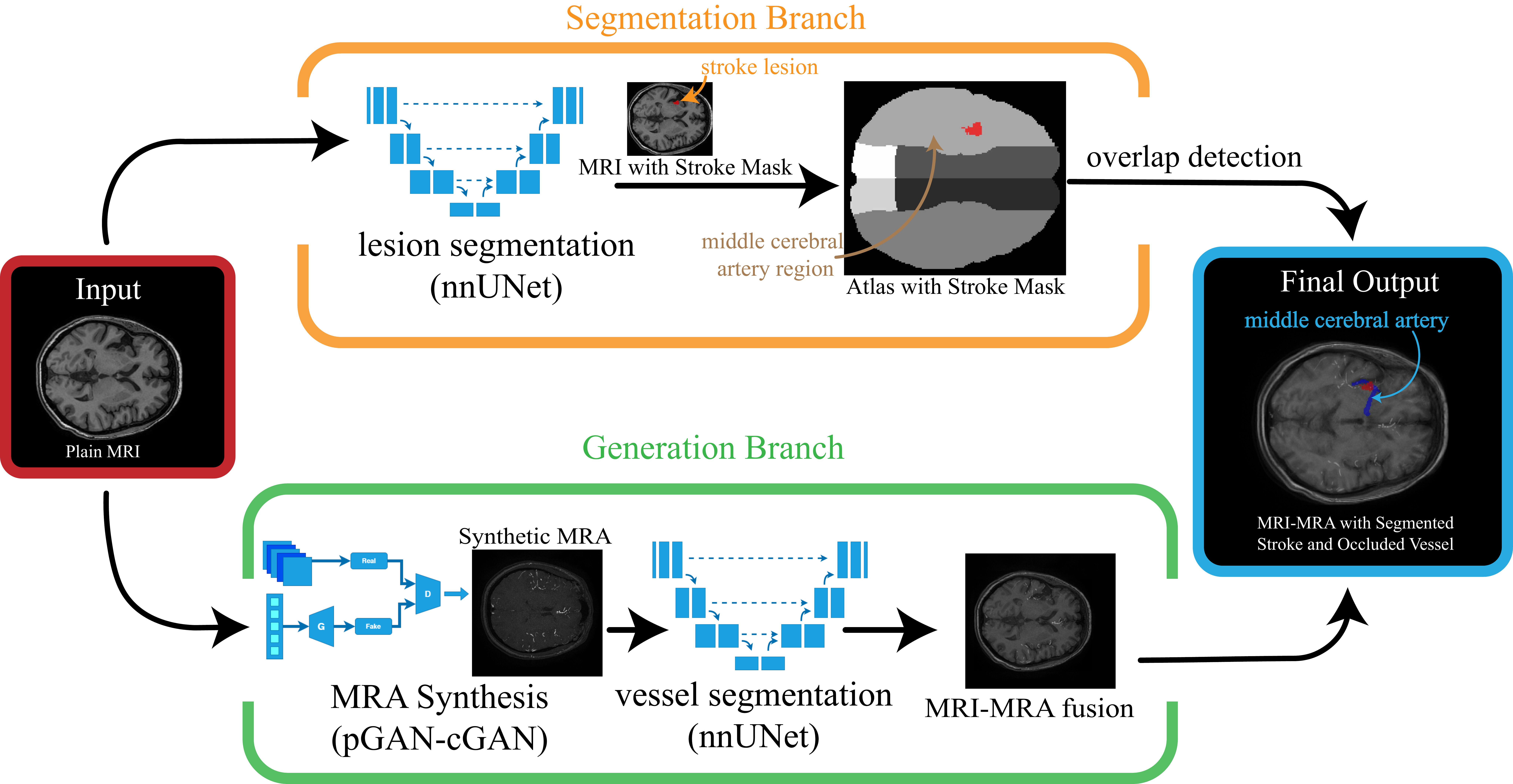}
    \caption{Stroke Locus Net Architecture Diagram}
    \label{fig:arch}
\end{figure}

Our system architecture consists of two branches, as shown in Fig. \ref{fig:arch}: the segmentation branch and the generation branch. The segmentation branch begins with lesion segmentation using nnU-Net as the backbone architecture. The output of this stage is the stroke lesion mask, which is then fed into the arterial probabilistic atlas component. This component determines the source of the stroke by analyzing the overlap between the lesion mask and predefined arterial territories.

In the generation branch, the input MRI passes through a GAN-based architecture to synthesize MRA images. The synthetic MRA is then used for vessel segmentation, allowing the identification and visualization of individual vessels. The MRI and MRA images are fused together to enhance visualization without the segmentation masks. Finally, the vessel class corresponding to the region identified as the source of the stroke is segmented and visualized using the vessel segmentation results. The final output shows both the stroke lesion mask and the mask corresponding to the source of the stroke.

\subsection{Segmentation Branch}
\subsubsection{Lesion Segmentation using nnUNet}
For lesion segmentation, we employ nnU-Net as our backbone architecture, leveraging its state-of-the-art performance in medical image segmentation tasks. We utilize the standard nnU-Net implementation without modifications, as its self-configuring nature automatically adapts to our specific dataset characteristics \cite{isensee2021nnunet,soliman2024deep}. The model is trained on the ATLAS v2.0 dataset \cite{liew2022atlas}, from which we use the T2-weighted MRI scans with manually annotated stroke lesion masks. During training, nnU-Net automatically determines optimal preprocessing steps, network architecture parameters, and training configurations based on the dataset properties. The model outputs binary segmentation masks indicating the presence of stroke lesions, which are then used as input for the arterial territory analysis component.

\subsubsection{Arterial Atlas Analysis}
We implement automated arterial territory analysis using the probabilistic arterial territory atlas developed by Liu et al. \cite{liu2023probabilistic}. This atlas subdivides the cerebral vasculature into ten distinct regions based on arterial supply: right and left middle cerebral artery territories (MCAR, MCAL), right and left anterior cerebral artery territories (ACAR, ACAL), right and left posterior cerebral artery territories (PCAR, PCAL), right and left vertebrobasilar regions (VBR, VBL), and right and left lateral ventricular regions (LVR, LVL). These classifications provide a fine-grained anatomical segmentation of the brain's vascular territories, offering region-specific associations between stroke lesions and their likely vascular origin.

The module accepts segmented stroke lesion masks and their corresponding MRI scans as inputs. Using FSL's FLIRT (FMRIB's Linear Image Registration Tool) \cite{jenkinson2001flirt1global,jenkinson2002flirt2improved}, we first perform rigid-body registration to align the patient's stroke mask with the arterial atlas space via a six-degree-of-freedom transformation. This transformation is then applied to the stroke mask. After registration, we calculate the voxel-wise overlap between the stroke lesion and each of the ten arterial territories. For each territory, the pipeline computes both the absolute number and the percentage of stroke voxels. The territory with the highest percentage overlap is identified as the dominant territory, suggesting the vascular source most likely responsible for the stroke.

\subsection{Generation Branch} 
\subsubsection{MRA Synthesis}
The generation branch begins with MRA synthesis using a pGAN (pixel-wise generative adversarial network) architecture as described in \cite{Dar2019MRISynthesis}. The pGAN consists of standard GAN blocks; the generator (G) and discriminator (D), as well as the front head of VGG16 until the second max pooling layer \cite{russakovsky2015imagenetlargescalevisual}. Statistical dependencies between the original MRI image and the generated MRA image are achieved by inputting the original image as the base for the transformation of the generator instead of random noise. This makes pGAN suitable for our "image-image" generation problem.

The loss function $L_{pGAN}$ consists of $L_{L1}$ loss (the pixel wise loss), $L_{perc}$ loss (the spacial features loss), and $L_{condGAN-k}(D,G)$ loss (the square loss of the adversarial network):

\begin{align*}
    L_{\text{condGAN-}k}(D,G) &= -\mathbb{E}_{\textbf{x}_k, y}\left[(D(\textbf{x}_k, y) - 1)^2\right] 
    - \mathbb{E}_{\textbf{x}_k}\left[(D(\textbf{x}_k, G(\textbf{x}_k)))^2\right]  \\
    L_{\text{L1-}k}(G) &= \mathbb{E}_{\textbf{x}_k, y}\left[\|y - G(\textbf{x}_k)\|_1\right]  \\
    L_{\text{perc-}k}(G) &= \mathbb{E}_{\textbf{x}_k, y}\left[\|V(y) - V(G(\textbf{x}_k))\|_1\right]  \\
    L_{\text{pGAN}} &= L_{\text{condGAN-}k}(D,G) + \lambda L_{\text{L1-}k}(G) + \lambda_{\text{perc}} L_{\text{perc-}k}(G) 
\end{align*}
Where $E$ is the expected output, $V$ is the output of VGG16 until the second maximum pooling layer, and $\lambda$ is the coefficient with which the pixel-wise loss and the spatial feature loss are scaled. This loss function encapsulates inter-slice dependencies and assumes perfect registration of the input and output images.

The training cycle involves the generator generating a fake MRA scan and the discriminator identifying if the scan is fake or from the set of real images in the training set. The objective of the discriminator is to maximize the loss function, while the generator must minimize it. After multiple iterations of training, the model stabilizes when the discriminator can no longer consistently distinguish fake scans from real scans.

The original implementation used MATLAB '.m' files as input, requiring an extensive preprocessing step before integration with typical works. Our modification involves the acceptance of NIFTI files as well as a preprocessing step where the MRA is registered to the MRI scan using metadata and resliced, thus making it compatible with our pipeline and maintaining the assumption that the input and expected images are aligned.

\subsubsection{Vessel Segmentation}
Building on the arterial territory analysis, we employ the vessel segmentation framework developed by Sabrowsky et al. \cite{Sabrowsky2023segmentation} to segment and localize the occluded vessel. This framework provides a pre-trained ensemble of models that classify cerebral vessels into four major classes: anterior circulation (including anterior cerebral arteries), posterior circulation (including posterior cerebral arteries), and left and right middle cerebral arteries.

However, the vessel segmentation framework's four-class scheme presents a coarser categorization compared to the ten-region classification used in the arterial atlas. To reconcile this mismatch in granularity, we developed a mapping strategy that consolidates the detailed arterial territories into broader vessel classes. Specifically, MCAR is mapped to the right middle cerebral artery class, MCAL to the left middle cerebral artery class, ACAR and ACAL to the anterior circulation class, and PCAR and PCAL to the posterior circulation class. The vertebrobasilar (VBR, VBL) and lateral ventricular (LVR, LVL) regions are excluded from this mapping, as they do not correspond to any of the vessel segmentation framework's four predefined classes and fall outside its classification scope.

This region-to-vessel consolidation enables seamless integration between the atlas-based dominant territory identification and the vessel segmentation output, ensuring a consistent and interpretable framework across both components of the pipeline.

\subsubsection{MRI-MRA Fusion}
Following the MRA generation, a fusion model takes the MRA and MRI images as inputs and fuses them using the CNN model \cite{zhang2009IFCNN}. The model consists of two convolutional layers with shared weights that fuse images using a pixel-wise function,  either the mean, sum, or max pooling of the pixels in the image. The result is passed through two more convolutional layers to reconstruct the fused image. For the first kernel, the weights are not trainable and are initialized to be the same as the first kernel of ResNet101.

Data loading for this model is done similarly to the data loading for MRI synthesis. The MRI and MRA volumes from our dataset are first aligned and registered, then they are adjacently re-sliced to provide the same slice for both MRA and MRI. Moreover, the input of the model is modified to accept NIFTI files.

\begin{figure}
    \centering
    \includegraphics[width=0.75\linewidth]{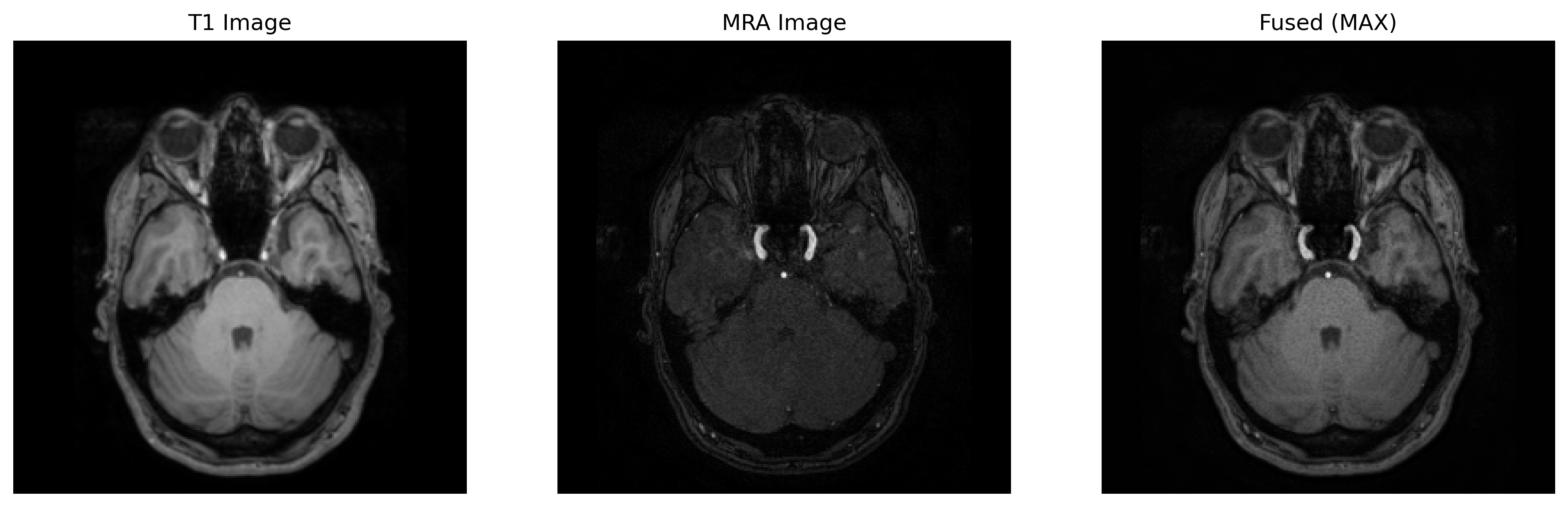}
    \includegraphics[width=0.75\linewidth]{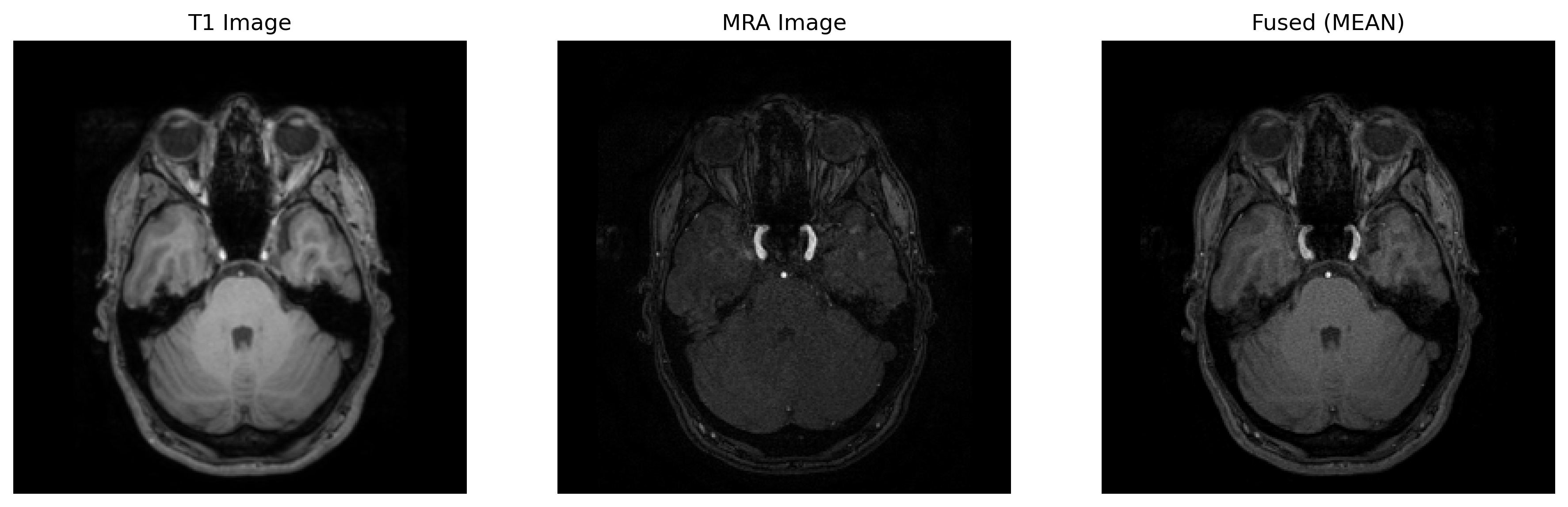}
    \includegraphics[width=0.75\linewidth]{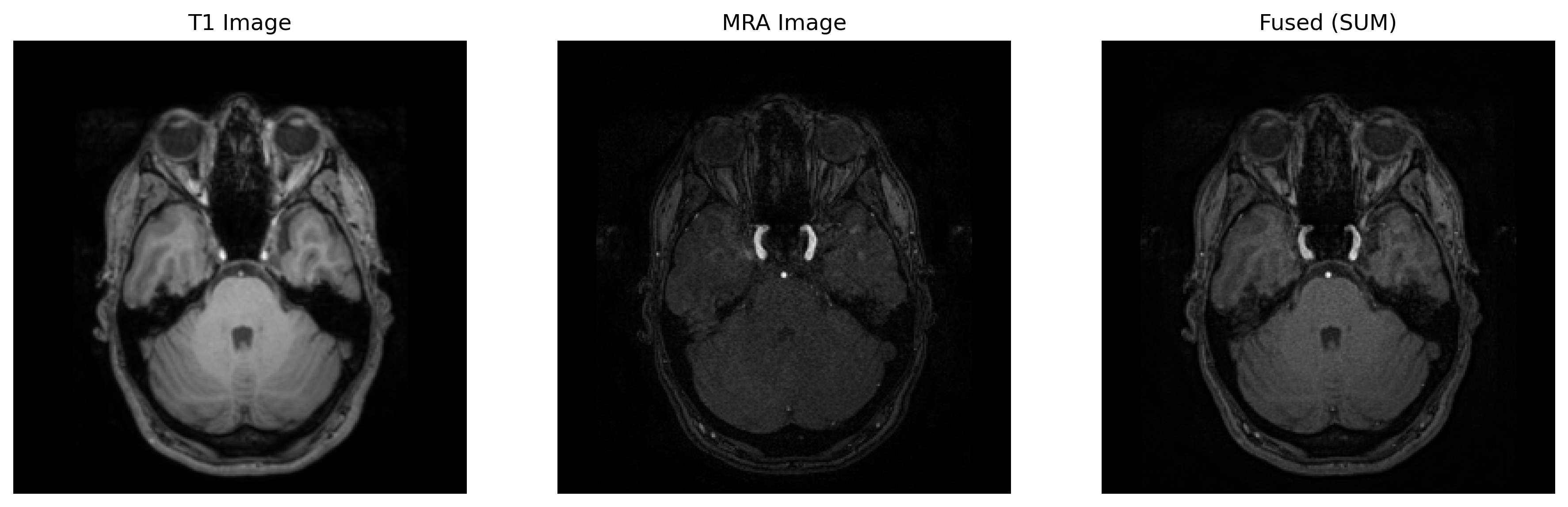}
    \caption{Fused image max (top), fused image mean (middle), fused image sum (bottom)}
    \label{fig:FusedImages}
\end{figure}

Following successful fusion, visualization is performed using all three aggregation modes. The results indicate that the mean and sum modes are the most effective, as they provide the highest contrast in vessel bright spots and body structures within the MRA images. This enhanced contrast makes the vessel structures more distinguishable while preserving the salient features of both the MRI and MRA images.

\section{Results and Discussion}

\begin{figure}
    \centering
    \includegraphics[width=0.8\linewidth]{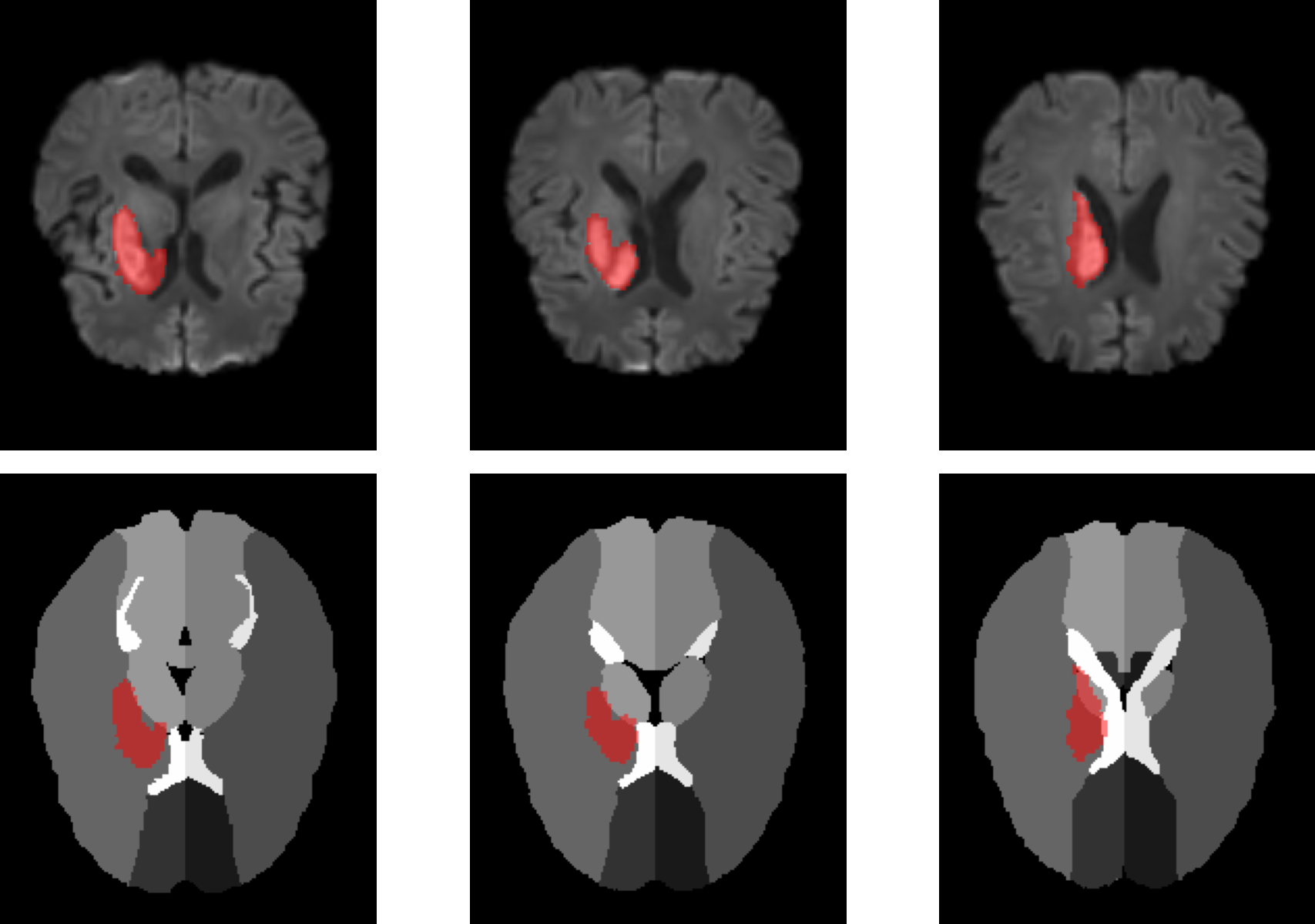}
    \caption{Multi-slice visualization of stroke lesion mask (red) overlaid on the arterial territory atlas and the MRI scan.}
    \label{fig:multi_slice}
\end{figure}

Figure \ref{fig:multi_slice} demonstrates the effectiveness of our arterial territory analysis module in localizing stroke lesions across multiple axial slices. The visualization shows how the stroke lesion intersects with different arterial territories defined by the atlas. The clear visualization of lesion-territory overlap enables clinicians to better understand the affected vascular regions and identify the occluded vessel responsible for the stroke. These territorial overlaps are used to determine the stroke source, which is used to determine which vessel is to be segmented in the vessel segmentation module.

\begin{figure}
    \centering
    \subfloat[Vessel segmentation using MRI]{
        \includegraphics[width=0.48\textwidth]{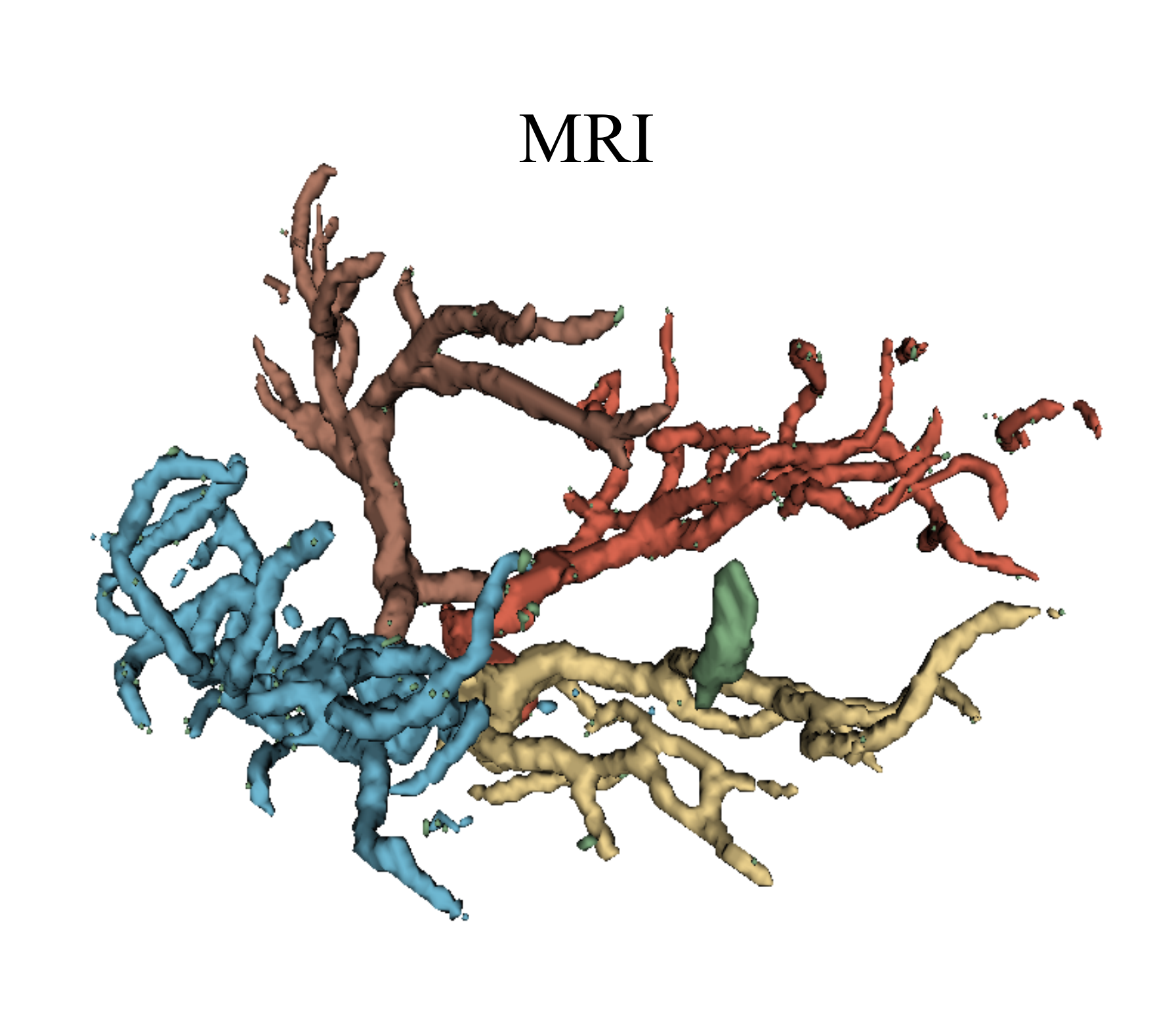}
        \label{fig:MRI}
    }
    \hfill
    \subfloat[Vessel segmentation using MRA]{
        \includegraphics[width=0.48\textwidth, trim=0 30 0 0, clip]{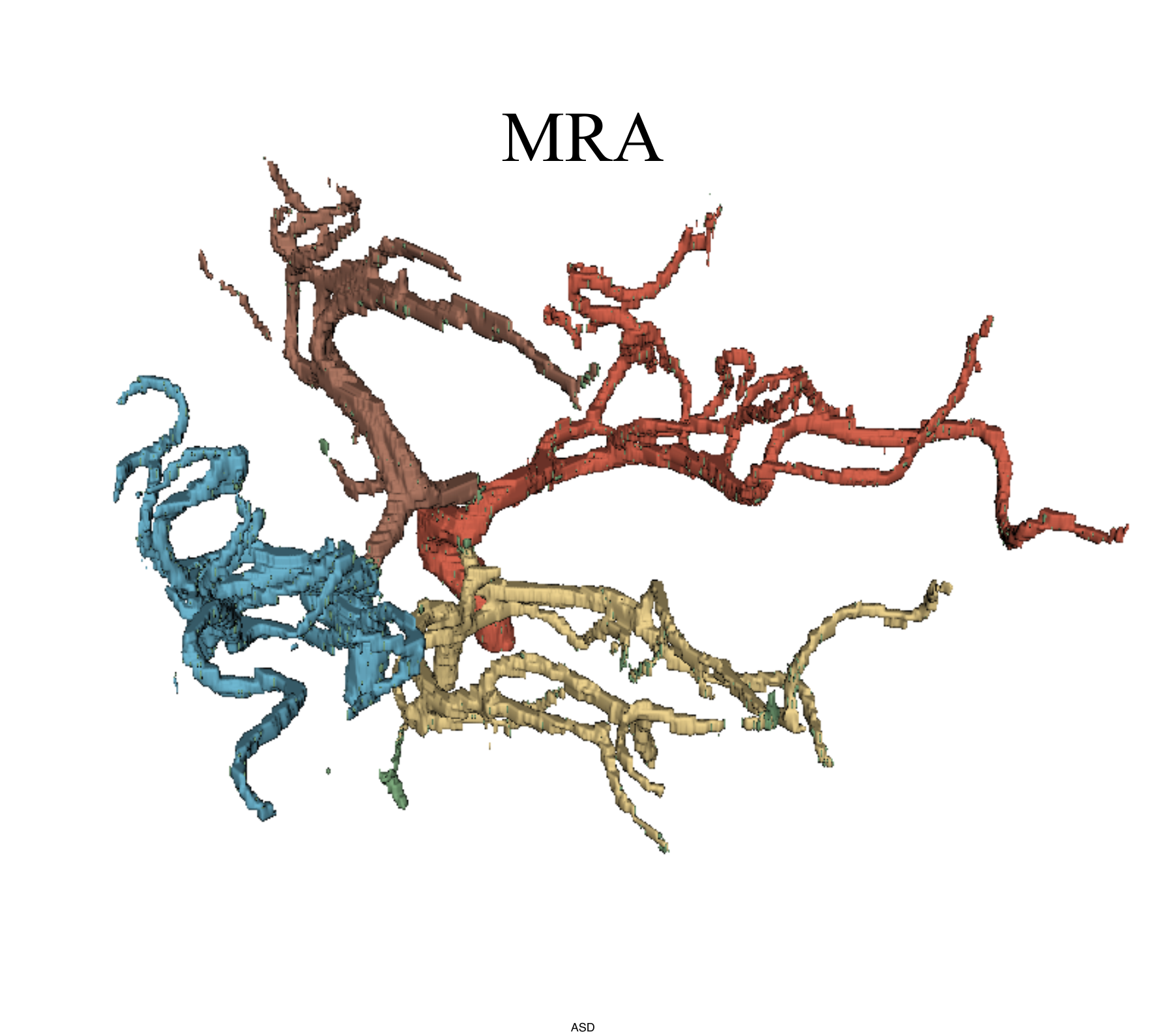}
        \label{fig:MRA}
    }
    \caption{3D visualization of quantitative vessel segmentation performance across modalities}
    \label{fig:MRIMRA}
\end{figure}

As shown in Figure \ref{fig:MRIMRA}, we implemented the vessel localization module \cite{Sabrowsky2023segmentation}, which demonstrated promising results in identifying the source of stroke lesions. Their work achieved a Dice Similarity Coefficient (DSC) of 0.84 for MRA-based vessel segmentation and 0.75 for T2 MRI-based segmentation. While we initially planned to feed our synthetic MRA images into their vessel localization module for our pipeline, the generated MRA quality was not sufficiently identical to the original MRA, as shown in Section \ref{sec:GeneratedMRAQuality}. Due to this, our current pipeline implementation uses MRI-based vessel localization instead of the originally intended MRA-based approach.

\label{sec:GeneratedMRAQuality}

\begin{figure}
    \centering
    \subfloat[Synthetic MRA]{
        \includegraphics[width=0.3\textwidth]{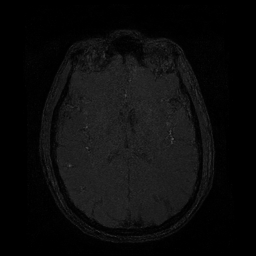}
        \label{fig:fake_mra}
    }
    \hfill
    \subfloat[Real MRA]{
        \includegraphics[width=0.3\textwidth]{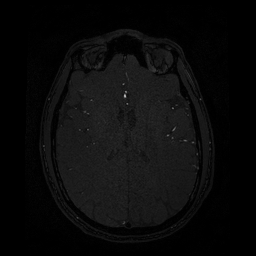}
        \label{fig:real_mra}
    }
    \hfill
    \subfloat[Real MRI]{
        \includegraphics[width=0.3\textwidth]{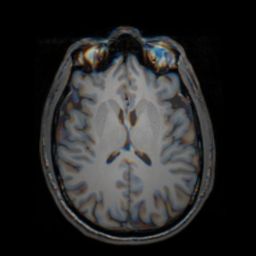}
        \label{fig:real_mri}
    }
    \caption{Comparison of synthetic MRA generation with ground truth MRA and input MRI}
    \label{fig:mra_comparison}
\end{figure}

For our MRA generation experiments, we utilized the IXI dataset \cite{ixi_dataset}. While the dataset provides high-quality paired MRI and MRA scans, it exclusively contains data from healthy subjects. This limitation is particularly relevant for our stroke-focused application, as the vascular patterns in healthy individuals may not fully represent those of stroke patients. This can only be mitigated by using a more diverse dataset that includes stroke patients, which is currently not available and beyond the scope of this work.

As shown in Figure \ref{fig:mra_comparison}, the MRA generation module successfully captures the overall structural information from the input MRI, but fails to generate the fine vascular details present in the real MRA. Thus, it lacks the distinct vessel patterns that are crucial for the subsequent vessel segmentation step in our pipeline. This limitation makes the generated MRA unsuitable for direct use in vessel localization, as the segmentation module requires clear vessel boundaries and contrast to accurately identify and classify vascular structures. With improvements to the MRA generation branch, connecting the two modules together results in a method with great potential for generating vessel segmentations through MRI alone.

\begin{table}[h]
    \centering
    \begin{tabular}{lcc}
        \toprule
        \textbf{Metric} & \textbf{Stroke Locus Net} & \textbf{pGAN-cGAN} \\
        \midrule
        PSNR & 17.82 $\pm$ 2.26 & 27.07 $\pm$ 0.62 \\
        SSIM & 0.31 $\pm$ 0.01 & 0.88 $\pm$ 0.03 \\
        \bottomrule
    \end{tabular}
    \caption{Comparison of image quality metrics between our T2 to MRA translation and the pGAN-cGAN T1 to T2 translation}
    \label{tab:metrics_comparison}
\end{table}

Table \ref{tab:metrics_comparison} shows a comparison of our image quality metrics with those reported in the pGAN-cGAN paper \cite{Dar2019MRISynthesis}. Notably, our implementation differs in target modalities; we translate between different imaging techniques (T2 MRI to MRA), while pGAN-cGAN translates between different MRI contrasts (T1 to T2). The lower PSNR and SSIM scores in our approach can be attributed to the greater complexity of generating vascular structures in MRA from T2 MRI, compared to the relatively simpler task of contrast translation between T1 and T2 MRI.

\section{Conclusion}
This study introduces Stroke Locus Net, a novel deep learning pipeline that localizes occluded vessels using only MRI images. By combining lesion segmentation, atlas-based vessel localization, MRI-to-MRA synthesis, and vessel segmentation, our framework offers a comprehensive approach to stroke diagnosis without relying on high-resolution vascular imaging. A combination of our atlas model output, our vessel generation output, our MRI and MRA fusion, with the stroke lesion mask, provides all the necessary information and visualization for a medical doctor to accurately determine the responsible vessel as well as providing early insight and diagnosis to aid the doctor's investigation and saving time.

The pipeline demonstrates the feasibility of using probabilistic atlases to determine the stroke's vascular source in its segmentation branch. In the generation branch, we demonstrate the feasibility of MRA generation from MRI images using the pGAN MRA synthesis module. Separately, vessel segmentation from both MRI and MRA images is demonstrated. Finally, the fusion of the MRI and the MRA images is shown to be effective in aiding the visualization of vascular structures in MRI images. 

However, challenges remain in the integration of the entire system as well as the effectiveness of the pGAN model in the generation of MRA images. Future work will focus on integrating all the modules into one seamless dataflow as well as enhancing the generation of vascular details through architectural improvements and more sophisticated training strategies to enable seamless integration with the vessel segmentation module. Initial ideas suggest changing hyperparameters, such as learning rate or increasing the coefficient of the pixel-to-pixel loss in the objective function of the pGAN model. This ensures the small pixel-to-pixel loss affects gradient descent more significantly. Further ideas suggest experimentation with MRA ground truth preprocessing to reveal salient features, easing vessel identification. Additionally, we urge the medical community to create more diverse publicly available datasets, especially paired MRI and MRA scans of stroke patients. Nevertheless, this work lays essential groundwork for future systems that can accelerate stroke diagnosis and improve accessibility in clinical settings, where time, resources or data is limited.

\section*{Acknowledgment}
We gratefully acknowledge the financial support provided by the Qatar National Research Fund (QNRF), a member of the Qatar Research, Development, and Innovation (QRDI) Council, through the Undergraduate Research Experience Program (UREP) under grant number UREP31-077-2-025.

\bibliographystyle{splncs04}
\bibliography{references}
\end{document}